\documentclass[sigconf]{acmart}

\usepackage{graphicx}
\usepackage{amsmath}
\usepackage{enumitem}
\usepackage{array}
\usepackage{listings}
\usepackage{multirow}
\usepackage[ruled, noend, linesnumbered]{algorithm2e} %
\usepackage{graphicx}

\newcolumntype{P}[1]{>{\centering\arraybackslash}p{#1}}
\newcolumntype{M}[1]{>{\centering\arraybackslash}m{#1}}

\newcommand{\onto}[1]{\textit{\textsf{\small{#1}}}}

\AtBeginDocument{%
  \providecommand\BibTeX{{%
    \normalfont B\kern-0.5em{\scshape i\kern-0.25em b}\kern-0.8em\TeX}}}

\copyrightyear{2021}
\acmYear{2021}
\setcopyright{acmcopyright}\acmConference[IJCKG'21]{The 10th
International Joint Conference on Knowledge Graphs}{December 6--8,
2021}{Virtual Event, Thailand}
\acmBooktitle{The 10th International Joint Conference on Knowledge
Graphs (IJCKG'21), December 6--8, 2021, Virtual Event, Thailand}
\acmPrice{15.00}
\acmDOI{10.1145/3502223.3502243}
\acmISBN{978-1-4503-9565-6/21/12}

\begin{document}


\title[Towards Ontology Reshaping for KG Generation: Applied to Bosch Welding]{Towards Ontology Reshaping for KG Generation \\ with User-in-the-Loop: Applied to Bosch Welding}


\author{Dongzhuoran Zhou}
\authornote{Dongzhuoran Zhou and Baifan Zhou contributed equally to this work as first authors.}
\email{dongzhuoran.zhou@de.bosch.com}
\affiliation{%
  \institution{Bosch Center for AI, Germany}
  \institution{SIRIUS Centre, University of Oslo}
  \city{}
  \country{}
}

\author{Baifan Zhou}
\authornotemark[1]
\email{baifanz@ifi.uio.no}
\affiliation{%
  \institution{SIRIUS Centre, University of Oslo}
  \city{Oslo}
  \country{Norway}
}

\author{Jieying Chen}
\email{jieyingc@ifi.uio.no}
\affiliation{%
  \institution{SIRIUS Centre, University of Oslo}
  \city{Oslo}
  \country{Norway}
}

\author{Gong Cheng}
\email{gcheng@nju.edu.cn}
\affiliation{%
  \institution{Nanjing University, China}
  \city{Nanjing}
  \country{China}
}

\author{Egor V. Kostylev}
\email{egork@ifi.uio.no}
\affiliation{%
  \institution{Department of Informatics}
  \institution{University of Oslo, NO}
  \city{}
  \country{}
}

\author{Evgeny Kharlamov}
\email{evgeny.kharlamov@de.bosch.com}
\affiliation{%
  \institution{Bosch Center for AI, Germany}
  \institution{SIRIUS Centre, University of Oslo}
  \city{}
  \country{}
}

\renewcommand{\shortauthors}{Zhou, et al.}

\begin{abstract}

Knowledge graphs (KG) are used in a wide range of applications. The automation of KG generation is very desired due to the data volume and variety in industries. One important approach of KG generation is to map the raw data to a given KG schema, namely a domain ontology, and construct the entities and properties according to the ontology. However, the automatic generation of such ontology is demanding and existing solutions are often not satisfactory. An important challenge is a trade-off between two principles of ontology engineering: knowledge-orientation and data-orientation. The former one prescribes that an ontology should model the general knowledge of a domain, while the latter one emphasises on reflecting the data specificities to ensure good usability. We address this challenge by our method of ontology reshaping, which automates the process of converting a given domain ontology to a smaller ontology that serves as the KG schema. The domain ontology can be designed to be knowledge-oriented and the KG schema covers the data specificities. In addition, our approach allows the option of including user preferences in the loop. We demonstrate our on-going research on ontology reshaping and present an evaluation using real industrial data, with promising results.

\end{abstract}



\maketitle
\section{Introduction}
\label{sec:intro}

Knowledge graphs (KGs) allow to structure information in terms of nodes and edges, 
where the nodes represent entities, 
e.g., welding machines, 
and edges connect entities and  thus represent relationships between them, e.g., by assigning software systems to concrete welding machines, or edges connect entities to their data values,
e.g., by assigning the weight and price to welding machines~\cite{hogan2021knowledge}.
In the context of Industry 4.0~\cite{kagermann2015change} and Internet of Things~\cite{mcewen2013designing}, KGs have been successfully used in a wide range of applications and industrial sectors in well known production companies such as Bosch~\cite{zhou2020cikm,DBLP:conf/semweb/KalayciGLXMKC20,DBLP:conf/semweb/StrotgenTFMTMA019}, Siemens~\cite{hubauer2018use,DBLP:conf/semweb/KharlamovSOZHLRSW14,DBLP:journals/ws/KharlamovMMNORS17}, Festo~\cite{DBLP:conf/semweb/ElmerJLMOSW17}, Equinor~\cite{DBLP:conf/semweb/SkjaevelandLH13,DBLP:conf/semweb/KharlamovHJLLPR15,DBLP:journals/ws/KharlamovHSBJXS17}, etc.\looseness=-1 


An important challenge in scaling the use of KGs in industry is to facilitate automation of KG construction from industrial data
due to its complexity and variety. 
Indeed, a common approach on KG construction is to construct entities and properties by relying on a given knowledge graph schema or ontology.
The KG schema defines upper-level concepts of data, and consists of classes and properties. 
A classical domain ontology is a formal specification of shared conceptualisation of knowledge~\cite{smith2012ontology, guarino2009ontology} and it reflects  experts knowledge on upper level concepts, specific domains, or applications. \looseness=-1

However, since the domain ontologies are knowledge-oriented, they do not focus on the specificities of arbitrary datasets.
In industry, data often come with a wide range of varieties. Many attributes exist in some datasets but not in others, and many terms in the domain ontologies do not exist in all datasets.
Thus, directly using domain ontologies as KG schemata
can naturally lead to a number of issues. Indeed, the resulting KG can contain a high number of blank-nodes, it may be affected by information loss, incomplete data coverage, and happen to be not user-friendly enough for applications.\looseness=-01

\begin{figure}[t]
\vspace{-1ex}
\includegraphics[width=.40\textwidth]{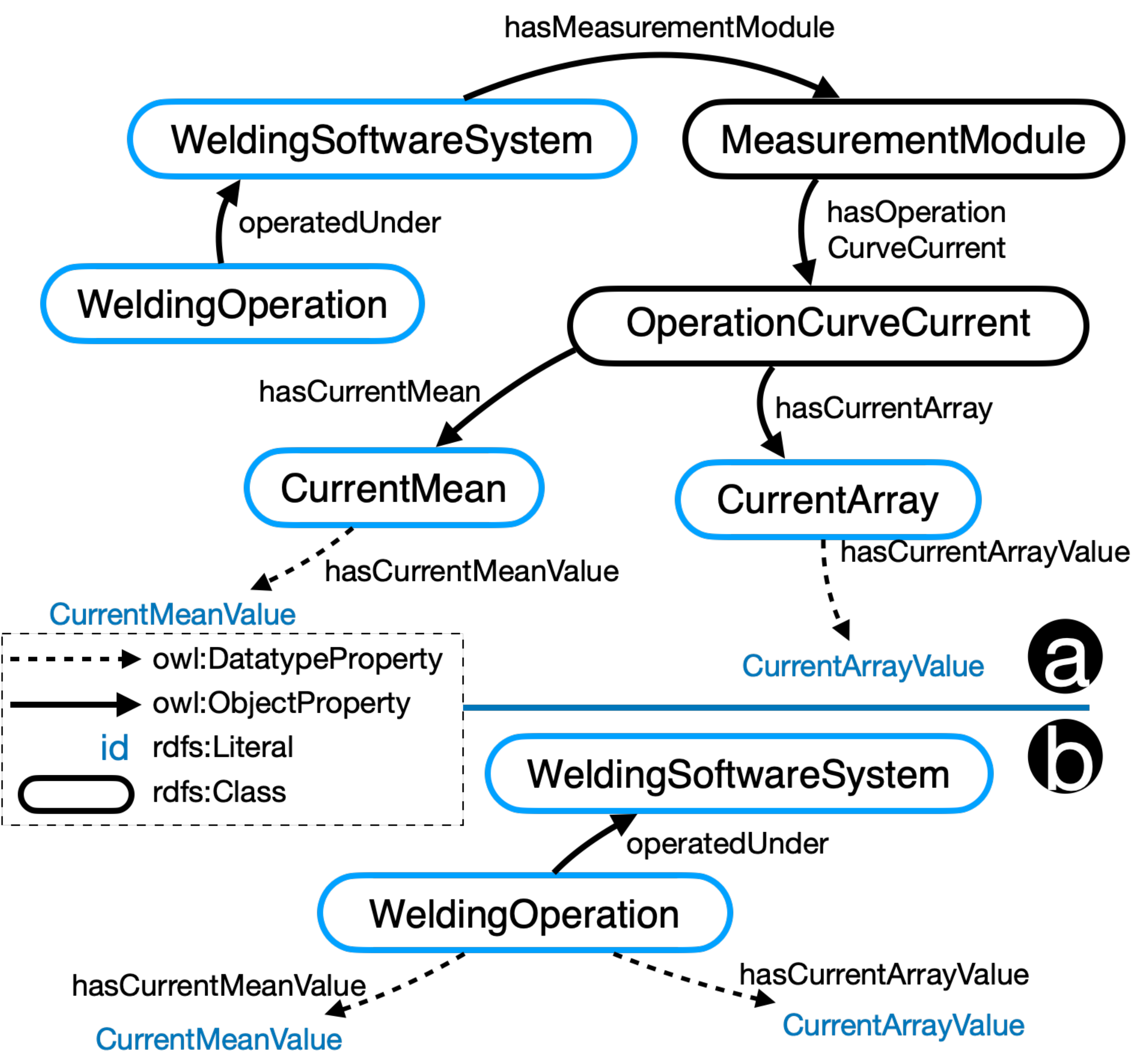}
 \vspace{-3ex}
\caption{\small{Why do we need ontology reshaping? Domain ontology (partially shown in a) reflects the knowledge; the KG schema (partially shown in b) needs to reflect raw data specificities and usability. Blue boxes: classes that can be mapped to attributes in the raw data; black boxes: classes that cannot be found in the raw data. \looseness=-1}}
\label{fig:intro-why-reshaping}
\vspace{-8ex}
\end{figure}

Considering an example in automated welding~\cite{zhouecda2018,zhou2021method}, which is an essential manufacturing process for producing the hundreds of thousands of car bodies in car factories everyday.
The domain ontology (Figure~\ref{fig:intro-why-reshaping}a) shows that, in the welding process, the welding operations are operated under welding software systems, which have measurement modules. These modules measure a sensor signal and save it as an operation curve, namely the current curve. The current curve is stored as an array, and it has a mean value. This presentation is created in close collaboration with the domain (welding) experts. It is thus intuitive  for human to understand the domain knowledge well.
Yet, the real datasets differ from the mental model.
In most welding datasets, there exist one current mean value and one current array value for each welding operation (highlighted with blue boxes), while one welding software system is responsible for a huge group of welding operations, and there exist no attribute for the measurement module and operation curve current.
If the domain ontology in Figure~\ref{fig:intro-why-reshaping}a is used for KG generaton directly, the KG will contain a lot of blank nodes or dummy nodes generated by the class \onto{MeasurementModule} and \onto{OperationCurveCurrent}. Furthermore, the user will not be able to find the one-to-one correspondences between \onto{WeldingOperation}, \onto{CurrentMeanValue} and \onto{CurrentArrayValue}. Meanwhile, the deep structure of the domain ontology will make the applications over the KG (e.g. query-based analytics) excessively complicated. Instead, the ideal schema for the KG generation would be Figure~\ref{fig:intro-why-reshaping}b, in which the ontology is significantly simplified.
The generated KGs will have zero dummy nodes,   and a 
much simpler structure. The KGs will be more efficient to generate, and easier to understand and use for the users.

The above-mentioned example exhibits 
the challenge of a trade-off between two principles of ontology engineering, (1) knowledge-orientation, which reflects the generality of domains, and (2) data-orientation, which aims at  KG generation well-suited for specificities of data. The former principle focuses on conveying the meaning, or knowledge of a given domain, while the latter principle emphasises on the usability of ontologies for applications~\cite{jarrar2005towards}.

To address this challenge, we propose our  \textit{ontology reshaping} method that 
computes KG schemata  from  domain ontologies; these schemata  in turn allow for generation of KGs of high quality.
In this way, we circumvent the trade-off problem and can use both ontologies for their different purposes.
Our contributions are as follows:\looseness=-1

\begin{itemize}[topsep=3pt,parsep=0pt,partopsep=0pt,itemsep=0pt,leftmargin=*]
     \item We propose an algorithm for ontology reshaping, which converts domain ontologies that reflect general knowledge to smaller KG schemata that cover data specificities, addressing the issue of sparse KG with dummy nodes.
    \item We design and conduct experiments for a proof-of-concept evaluation. We derive requirements for the use case, and design performance metrics for ontology reshaping.
\end{itemize}




\section{Preliminaries}
\label{sec:preliminaries}
%

\smallskip
\noindent
\textbf{Problem Formulation.}
Intuitively, ontology reshaping is about computing a smaller ontology from a larger one by taking a subset of its classes and possibly redefining some of its axioms based on some external heuristics as well as on some notion of optimality.
In this work we do not aim at developing a formal theory of ontology reshaping, but rather at providing intuitions behind this problem and preliminary solutions that account for a particular type of reshaping.
More precisely, in this work we consider reshaping where (i) re-definition of axioms is essentially ``re-assigning'' of properties from some classes to another  (ii) the external heuristics is user's input and (iii) the notion of optimality is the coverage of the data which the re-shaped ontology should be mapped to. 
In other words, such reshaping can be seen as: 




\vspace{-2ex} 
\begin{equation*}
     \text{S} \leftarrow f(O, D, M, MC, I_U)
\end{equation*}
\vspace{-3ex}

\noindent where $f$ is an algorithm that takes in the inputs and outputs a KG schema S. The inputs are: $O$ is a larger ontology, $D$ is the raw data, that $O$ is related to. The raw data $D$ is in the form of relational table. $M$ is a set of mappings that relate the attributes and table names in raw data to the classes in $O$. Apart from that, we need some more information given by the users. In our case, this includes two parts, (1) the users need to point out the most important entity (the most important class in $O$), named as the main class (MC); (2) User information $I_U$ some more optional information.



\smallskip
\noindent
\textbf{Requirements of KG.}
We derive the requirements as follows: 
\begin{itemize}[topsep=3pt,parsep=0pt,partopsep=0pt,itemsep=0pt,leftmargin=*]
    \item \textit{R1 Completeness.} The knowledge graphs should be able to completely represent the raw data.
    \item \textit{R2 Efficiency.} The generation of KG schema and KG should be efficient both in computational time and storage space.
    \item \textit{R3 Simplicity.} The KG schema should not be over-complicated for understanding and use: (1) The generated KG should not have too much redundant information, e.g. dummy entities that are generated solely because of the schema but have no correspondence in the raw data; (2) the KG should  not have complicated structures.\looseness=-1
\end{itemize}


%

\section{Our method: ontology reshaping}
\label{sec:solution}


\smallskip




\smallskip
\noindent \textbf{Algorithm Explanation.} Intuition of the algorithm in five steps:
\begin{itemize}[topsep=3pt,parsep=0pt,partopsep=0pt,itemsep=0pt,leftmargin=*]
    \item Step 1: Initialise the KG schema S with MC, and two sets from classes in $O$: potential classes $C_E$ and potential properties $C_P$.
    \item Step 2:  Add classes in $C_E$  to S if they are mapped to  table names T in $M$.
    \item Step 3: If properties in $C_P$ are named as 'ID', 'NAME' etc, add their corresponding classes from $C_E$ to S. Add further classes from $C_E$ to S according to user information $I_U$ (optional).
    \item Step 4: Connect classes in S and connect properties to classes according to $O$.
    \item Step 5: Connect the rest classes in S to MC or according to user information $I_U$ (optional)

\end{itemize}


\smallskip
\noindent \textbf{Step 1. Initialisation.} We start our  algorithm with initialisation (Line~1). S is initialised with the main class MC. Then we map classes in $O$ to attributes in $D$ with $M$, where classes that can be mapped to attributes in $M$ are initialised in the set of potential properties $C_P$, the rest classes are initialised as potential classes $C_E$. \looseness=-1


\smallskip

\noindent \textbf{Step 2. Class addition with table names.} The raw data D is in form of relational table. If the relational table's name is able to be mapped from $M$ to classes in $C_E$, We then add the mapped classes from $C_E$ to S (Line 2 to 3).


\smallskip
\noindent \textbf{Step 3. Entity identification by key words.}
In Step 3 we identify the classes in potential properties $C_E$ by key words and (optional) user information. We map the attributes A stored in raw data from $M$ to properties in $C_P$. Indeed, if properties in $C_P$ are named as 'ID', 'NAME', e.g., WeldingProgramID, WeldingMachineName, we then add the corresponding classes in $C_E$, e.g., WeldingProgram, WeldingMachine, to S. Besides, we add the specidied entities in S based on user information (Line 4 to 12).


\begin{algorithm}[t!]
\SetAlgoLined
\caption{Ontology Reshaping}
\label{algo:Ontoagg}
\KwIn{O, M, MC, D, $I_U$}
\KwOut{S} 
\SetKwProg{Init}{Initialisation: }{}{}
\vspace{1ex}
\text{S} $\leftarrow$\{MC\}, 
$\{C_E, C_P\} \leftarrow O$,
A, T $\leftarrow$ D \\ 
$C_{ET}\leftarrow$ (M, T) \\
S:= S $\cup$  \{$C_{ET}$\} \\ 
\ForEach{$A_{i} \in A$}{
$C_{Pi}\leftarrow(M,A_i)$ \\
\uIf{$I_U$ exists}{
$C_{Ei}\leftarrow$ClassIdentification($C_{Pi}$, $I_U$)\\
S:= S~$\cup~\{C_{Ei}$\} \\
}\uElseIf {'ID' or 'Name' in $C_{Pi}$}{
$C_{Ei} \leftarrow C_{Pi}$\\
S:= S~$\cup ~ \{C_{Ei}$\} \\
}
}
S$\leftarrow$ ClassConnection(S, MC, $O$, $I_U$) \\
\end{algorithm}
\setlength\abovedisplayshortskip{-6pt}
\setlength\belowdisplayshortskip{0pt}


\smallskip

\smallskip
\noindent \textbf{Step 4. Classes connection .}
In Step 4 (in algorithm \ref{algo:EntityClassCon}) we connect classes in S and connect properties to classes according to $O$. It takes 4 inputs:  KG schema S, Domain ontology O, main class MC, and user information $I_U$. We start the Algorithm \ref{algo:EntityClassCon} with initialisation (Line~1), where the $C_{rest}$ is initialised with copying classes in S . Then we iterate all permutation of the classes pairs ($C_i,C_j$) in S  (Line~2). For each pair ($C_i,C_j$), if the relation $r_{o}(C_i,C_j)$ already exists in S,  we then continue to the next pair (Line~3 to 4). Otherwise, if direct relation $r_{od}(C_i,C_j)$ exists in O but not in S (Line~5 to 6), then this relation will be added to S. 
In case of existing in the ontology an indirect relation between the two classes $C_i,~C_j$, the algorithm first evaluates if user information $I_U$ exists and adds a relation between classes $r_{o}(C_i,C_j)$ (Line 7 to 10), and if not both classes are related to the main class $r_{o}(MC,C_i)$,  $r_{o}(MC,C_j)$ (Line 11 to 12). \looseness=-1

\smallskip
\noindent \textbf{Step 5. Classes connection with UserInfo.} In Step 5 (in algorithm \ref{algo:EntityClassCon}) we connect classes in S  to MC  that are not connected to any classes, or according to UserInfo $I_U$ (optional) (Line 18 to 25).\looseness=-1

\section{Evaluation }
\label{sec:application}
\vspace{-1ex}


\smallskip
\noindent \textbf{Data Description.}
We have five inputs: domain ontology O, dataset D,  mappings M, main class MC and user information \textit{UserInfo}. \looseness=-1

The domain ontology O is an OWL 2 ontology and can be expressed in Description Logics $\mathcal{SHI(D)}$. With its 1249 axioms, which contain 147 classes, 145 object properties, and 132 datatype properties, it models the general knowledge of a type of fully automated welding process, resistance spot welding~\cite{svetashova2020,zhou2020cikm,ijckg2021mlkg,zhou2021jim},

The industrial dataset $D$ is collected from welding production lines in a factory in Germany. The raw form of $D$ is various, including txt, csv, RUI (a special format of Bosch time series data), SQL database, etc. They are transformed into relational tables. We selected a subset from the huge amount of data to evaluate our method. The transformed data contain one huge table of welding operation records and a series of tables of welding sensor measurements. In total, there exist about 4.315 million records. These data account for 1000 welding operations, estimated to be related to 100 cars.\looseness=-1

The mapping M consists of three mappings: (1) a meta mapping that gives the correspondence between the table names in $D$ and the class names in O; (2) an operation mapping that relates the attribute names of the welding operation records to the classes in O; (3) a sensor mapping that annotates the attribute names of the welding sensor measurements with the classes in O.

The UserInfo contains two parts, (1) mandatory: the users need to point out the main entity ME, which is the welding operation; (2) optional: other information, e.g. the attribute names corresponding to other possible entity names  and their related properties

\begin{algorithm}[t!]
\SetAlgoLined
\caption{Class Connection}

\label{algo:EntityClassCon}
\KwIn{S, MC, O, $I_U$}
\KwOut{S}
$C_{rest} \leftarrow$ ClassExtraction(S) \\ 
\ForEach{$ (C_i,C_j) \subseteq  S$ }{
    \uIf{$r_{o}(C_i,C_j) \in$ S}{ Continue 
    }
    \uElseIf{$r_{o}(C_i,C_j) \in$ O }    
    {S:=S $\cup \{ r_{o}(C_i,C_j)\}$ 
    }
    \uElseIf{$ r_{oind}(C_i,C_j) \in$ O}{ 
    \eIf{$I_U$ exists}{
    \{$r_o$\} $\leftarrow$ UserInfoExtraction($I_U$)\\
    S:=S $\cup \{r_{o}(C_i,C_j)$\} \\
    }{
    S:=S $\cup \{r_{o}(MC, C_i), r_{o}(MC, C_j)$ \}\\
    }
    }
}
$C_S \leftarrow $ClassExtraction(S)\\
$C_{rest} \leftarrow C_{rest} \setminus C_S$   \\
\ForEach{$ C_i \in  C_{rest}$ }{
    \eIf{$I_U$ exists}{
    \{$r_d(C_i,C_u),r_d(C_u,C_i)$\} $\leftarrow$ UserInfoExtraction($I_U$)\\
    S:=S $\cup$ \{$r_d(C_i,C_u), r_d(C_u,C_i)$\} \\
    }{
    S:=S $\cup \{ r_{d}(MC, C_i)\}$ \\
    }
}

\end{algorithm}

\smallskip
\noindent \textbf{Experiment Design.}
To test whether our ontology reshaping algorithm can perform well for an arbitrary dataset,
we randomly sub-sample the dataset $D$ to 6 sub-datasets (Set 1-6 in Table~\ref{table:DataMetrics}). Each set contains a subset of the attributes of D, reflecting different data complexity. 
The numbers of attributes in the subsets increase by ten each time, from 10 to 60. We repeat the sub-sampling for each subset 10 times to decrease the randomness, and provide the mean values of the evaluation metrics (introduced in next paragraph). 

We compare two approaches: the \textit{baseline} of KG generation without ontology reshaping; KG generation with our method of ontology reshaping.
The baseline is a naive approach to select a subset from the domain ontology $O$ and use it as the KG schema. The subset includes  (1) classes that have correspondence to attributes in raw data, and (2) the classes connect these classes in (1).

In total, we performed 120 experiments ($2 \times 6 \times 10$): 2 approaches (the baseline and KG generation with ontology reshaping), 6 times sub-sampling with 10 repetition for each.



\smallskip
\noindent \textbf{Evaluation Metrics.}
We use three sets of metrics to evaluate the fulfilment of the three requirements in Sect.~\ref{sec:preliminaries}: completeness, efficiency and simplicity. To evaluate \textit{completeness},
we have \textit{data coverage} that represents the percentage of attributes in raw data covered by the generated KG.\looseness=-1
 \textit{Efficiency}  measures the ability of the approaches to use the least time cost for generating KG from raw data and to take the least storage space and the least entities and properties to represent the same information of raw data. The efficiency metrics thus include time cost, storage space, number of classes in the KG schema, and number of entities and properties (including object properties and data properties) in the generated KG.
\textit{Simplicity} measures the performance of the approaches to represent the same raw data with the simplest KG. These metrics include the number of dummy entities in the generated KG (namely the entities that cannot be mapped to attributes in the raw data, but are generated solely because of the KG schema),  and two depths (the depth characterise the number of edges to connect two nodes in the KG via the shortest path): (1) root to leaf depth measures the ``depth'' to find the furthest entity starting from the main entity; (2) the global depth is the largest depth across the whole KG.

\begin{table}[t]
 
\renewcommand*{\arraystretch}{1.1}
\footnotesize
	\setlength{\tabcolsep}{0.55mm}
	\center
    \caption{\small{Evaluation on subsets with different numbers of attributes shows that KG generation with the ontology reshaping (Onto-Reshape) outperforms the baseline significantly in terms of efficiency and simplicity. The data coverage of both methods is 100\%, and is thus not displayed in the table. Avg.: average, prop.: property.}} 
    \label{table:DataMetrics}
    \vspace{-4ex}
        \begin{tabular}{M{0.92cm}|P{2.5cm}|P{0.68cm}|P{0.68cm}|P{0.68cm}|P{0.68cm}|P{0.68cm}|P{0.68cm}}
        \hline
        \multicolumn{2}{c|}{Subset} &Set 1&Set 2&Set 3&Set 4&Set 5&Set 6 \\\hline \hline
 \multirow{2}{=}{Raw data} & \#attributes & 10 & 20 & 30 & 40 & 50 & 60 \\ \cline{2-8}
 & storage space (kB) & 81.3 & 164.7 & 222.5 & 309.2 & 379.5 & 451.3 \\ \hline
 \multicolumn{8}{c}{Efficiency metrics}\\ \hline
\multirow{7}{=}{Baseline}  & time cost (sec) & 139.3 & 298.2 & 429.5 & 551.6 & 723.3 & 836.9  \\ \cline{2-8}
 & storage space (MB) & 1.3  & 2.0  & 2.7  & 3.4  & 4.1  & 4.8   \\ \cline{2-8}
 & \#avg. class & 147.0 & 147.0 & 147.0 & 147.0 & 147.0 & 147.0  \\ \cline{2-8}
 & \#max. class & 147.0 & 147.0 & 147.0 & 147.0 & 147.0 & 147.0 
  \\ \cline{2-8}
 & \#object prop. & 14.1k & 25.0k & 34.8k & 44.6k & 54.3k & 64.3k  \\ \cline{2-8}
 & \#data prop. & 404.5 & 577.8 & 1174.4 & 1546.9 & 2219.1 & 2100.4  \\ \cline{2-8}
 & \#entities &1704.5 & 1977.8 & 2474.4 & 3346.9 & 4319.1 & 4048.4  \\ \hline
\multirow{7}{=}{Onto-Reshape} & time cost (sec) & 21.6 & 40.8 & 61.2 & 84.9 & 113.3 & 128.0  \\ \cline{2-8}
 & storage space (MB) & 0.5  & 0.8  & 1.2  & 1.6  & 2.0  & 2.4   \\ \cline{2-8}
 & \#avg. class & 1.3 & 1.4 & 1.3 & 1.8 & 2.1 & 1.9  \\ \cline{2-8}
 & \#max. class & 2.0 & 2.0 & 3.0 & 3.0 & 3.0 & 3.0  \\ \cline{2-8}
 & \#avg. object prop. & 300.0 & 400.0 & 306.0 & 806.8 & 1120.4 & 913.6  \\ \cline{2-8}
 & \#avg. data prop. & 7.9k & 14.6k & 22.9k & 29.7k & 37.5k & 45.3k  \\ \cline{2-8}
 & \#entities & 1013.6 & 1020.4 & 1014.6 & 1028.2 & 1050.6 & 1029.2  \\ \hline
  \multicolumn{8}{c}{Simplicity metrics}\\ \hline
\multirow{6}{=}{Baseline}  & \#avg. dummy entities & 4100.0 & 4600.0 & 4800.0 & 4800.0 & 4600.0 & 4800.0 \\ \cline{2-8}
 & \#max. dummy entities & 5000.0 & 5000.0 & 6000.0 & 6000.0 & 6000.0 & 6000.0 \\ \cline{2-8}
 & avg. root to leaf depth & 3.2 & 3.3 & 3.3 & 3.3 & 3.4 & 3.3  \\ \cline{2-8}
&max. root to leaf depth & 4.0 & 4.0 & 4.0 & 4.0 & 4.0 & 4.0  \\ \cline{2-8}
& avg. global depth & 3.2 & 3.3 & 3.3 & 3.3 & 3.4 & 3.3  \\ \cline{2-8}
&max. global depth & 4.0 & 4.0 & 4.0 & 4.0 & 4.0 & 4.0  \\ \hline
\multirow{6}{=}{Onto-Reshape}  & \#avg. dummy entities & 0.0 & 0.0 & 0.0 & 0.0 & 0.0 & 0.0 \\ \cline{2-8}
 & \#max. dummy entities & 0.0 & 0.0 & 0.0 & 0.0 & 0.0 & 0.0 \\ \cline{2-8}

& avg. root to leaf depth & 1.3 & 1.4 & 1.2 & 1.7 & 2.1 & 1.7  \\ \cline{2-8}
& max. root to leaf depth & 2.0 & 2.0 & 2.0 & 2.0 & 3.0 & 2.0  \\ \cline{2-8}
& avg. global depth & 1.3 & 1.4 & 1.3 & 1.8 & 2.1 & 1.7  \\ \cline{2-8}
& max. global depth & 2.0 & 2.0 & 3.0 & 3.0 & 3.0 & 2.0  \\ \hline

        \end{tabular}
        \vspace{0ex}

    \vspace{-7.5ex}
\end{table}

\smallskip
\noindent \textbf{Results and Discussion.}
In Table~\ref{table:DataMetrics}, the six subsets with different number of randomly sub-sampled (each repeated 10 times) attributes  are listed as the columns. The data complexity increases from Set 1 to Set 6. Both approaches can represent the raw data 100\%. Thus it is not displayed in the table.\looseness=-1
We observe that OntoReshape outperforms the baseline in terms of time cost, storage space and other efficiency metrics, especially that the KG generation with OntoReshape is 7 to 8 times fast than that of the baseline.\looseness=-1
In terms of simplicity, OntoReshape also outperforms the baseline significantly.  The baseline generates a huge amount of dummy entities, while OntoReshape generates zero dummy entities, drastically reducing information redundancy. The two depths of the KGs generated by OntoReshape are also only half or one third of that of the baseline. The KGs generated by OntoReshape are thus much more simpler, and easier to understand for the users.
With an average root to leaf depth of only 1.2, we can see the users can use fewer queries to reach the deepest entities in the KG generated by OntoReshape.




\section{Related Work}

Knowledge graphs have received much attention in industries~\cite{zheng2021towards,hubauer2018use,ringsquandl2017event,garofalo2018leveraging}. KGs provide semantically structured information that can be interpreted by computing machines~\cite{zou2020survey,zhao2018architecture}, and an efficient foundation for standardised ways of data retrieval and analytics to support data driven methods. 
Data driven methods have been widely used in industries~\cite{zhou2017practical,zhou2019improving,mikhaylov2019high,mikhaylov2019machine}, especially machine learning~\cite{DBLP:conf/semweb/ZhouSPK20,DBLP:conf/semweb/SvetashovaZSPK20,DBLP:phd/dnb/Zhou21,zhou2020cikmdemo,zhou2021semml}. The problem of transforming a bigger ontology to a smaller ontology of the same domain is often referred to as ontology modularisation~\cite{conf/jelia/ChenLM019,conf/gcai/ChenL018,DBLP:conf/fois/ChenL016,DBLP:conf/semweb/ChenLMW17,ijcai2020-227} and ontology summarisation~\cite{ozacar2021karyon}. Most of them focus on the problem of selecting a subset of the ontology that is interesting for the users~\cite{konev2013model}, but they still cannot avoid dummy entities. Works on ontology reengineering~\cite{suarez2012introduction,suarez2012neon} also talked about reuse/adjustment of ontologies, but they do not focus on  automatically creating an ontology that reflect data specificities.\looseness=-1

\section{Conclusion
and
Outlook
}
\label{sec:conclusion}

This work presents our on-going research of ontology reshaping, that generates a small ontology covering data specificities
from a more complex domain ontology reflecting general knowledge.
The current approach can not fully retain the semantics of the domain ontology. This can be addressed by uniform interpolation, also known as forgetting~\cite{DBLP:conf/ijcai/KonevWW09, DBLP:conf/gcai/Ludwig016}, which we also plan to study in the future.
Furthermore,
we plan to compare our work with the body of research, where we actively contributed, on ontology evolution~\cite{DBLP:conf/kr/GrauJKZ12,DBLP:journals/ws/ZheleznyakovKNC19,DBLP:conf/aaai/ZheleznyakovKH17}, knowledge modelling and summarisation~\cite{DBLP:conf/semweb/KharlamovGJLMRN16a,DBLP:conf/www/0001GK20,DBLP:conf/esws/LiuCCKLQ20,XiaxiaISWC2021,XiaxiaTKDE2021,DBLP:conf/cikm/ChenWCKQ19,DBLP:conf/semweb/WangCLCPKQ19},
ontology extraction or bootstrapping~\cite{DBLP:conf/semweb/Jimenez-RuizKZH15,DBLP:journals/semweb/PinkelBJKMNBSST18},
and to investigate how to extend our work to account for ontology aggregation techniques~\cite{DBLP:conf/cikm/CalvaneseKNT08,DBLP:journals/ws/KharlamovKMNNOS19}, and to develop end-user interfaces for exploration and improvement of reshaped ontologies~
\cite{DBLP:conf/cikm/ArenasGKMZ14,DBLP:journals/ws/ArenasGKMZ16,DBLP:conf/cikm/KharlamovGSGKH17,DBLP:conf/semweb/SherkhonovGKK17,DBLP:journals/semweb/SoyluKZJGSHSBLH18,DBLP:journals/jaise/SoyluGSJKONB17,DBLP:journals/uais/SoyluGJKZH17}.\looseness=-1


\everypar{\looseness=-1}

\small
\noindent\textbf{Acknowledgements.}
The work was partially supported by the H2020 projects Dome 4.0 (Grant Agreement No. 953163), OntoCommons (Grant Agreement No. 958371), and DataCloud (Grant Agreement No. 101016835) and the SIRIUS Centre, Norwegian Research Council project number 237898. 

\bibliographystyle{ACM-Reference-Format}
\bibliography{2021ijckg}

\end{document}